\documentclass[10pt,twocolumn,letterpaper]{article}

\usepackage[pagenumbers]{cvpr} 

\usepackage{graphicx}
\usepackage{amsmath}
\usepackage{amssymb}
\usepackage{booktabs}

\usepackage{algorithm}
\usepackage{algorithmic}

\usepackage{multirow}
\usepackage{array} 

\usepackage[pagebackref,breaklinks,colorlinks]{hyperref}

\usepackage[capitalize]{cleveref}
\crefname{section}{Sec.}{Secs.}
\Crefname{section}{Section}{Sections}
\Crefname{table}{Table}{Tables}
\crefname{table}{Tab.}{Tabs.}

\def\cvprPaperID{*****} 
\def\confName{CVPR}
\def\confYear{2022}

\begin{document}

\title{Adversarial Learning for Neural PDE Solvers with Sparse Data}

\author{%
    Yunpeng Gong \\
    School of Informatics\\
    Xiamen University\\
    \texttt{fmonkey625@gmail.com, gongyunpeng@stu.xmu.edu.cn} \\
    \and
    Yongjie Hou \\  
    School of Informatics\\
    Xiamen University\\
    \texttt{23120231150268@stu.xmu.edu.cn} \\
    \and
    Zhenzhong Wang \\
    Department of Computing\\
    The Hong Kong Polytechnic University\\
    \texttt{zhenzhong16.wang@connect.polyu.hk} \\
    \and
    Zexin Lin \\
    School of Informatics\\
    Xiamen University\\
    \texttt{zexinlin@stu.xmu.edu.cn} \\
    \and
    Min Jiang\thanks{Corresponding author} \\
    School of Informatics\\
    Xiamen University\\
    \texttt{minjiang@xmu.edu.cn} \\
}

\maketitle

\begin{abstract}
Neural network solvers for partial differential equations (PDEs) have made significant progress, yet they continue to face challenges related to data scarcity and model robustness. Traditional data augmentation methods, which leverage symmetry or invariance, impose strong assumptions on physical systems that often do not hold in dynamic and complex real-world applications. To address this research gap, this study introduces a universal learning strategy for neural network PDEs, named Systematic Model Augmentation for Robust Training (SMART). By focusing on challenging and improving the model's weaknesses, SMART reduces generalization error during training under data-scarce conditions, leading to significant improvements in prediction accuracy across various PDE scenarios. The effectiveness of the proposed method is demonstrated through both theoretical analysis and extensive experimentation. The code will be available.
\end{abstract}

\section{Introduction}
Partial differential equations (PDEs) have a long-standing history of application across science and engineering, providing a formal mathematical framework to describe and solve dynamic systems involving multiple variables. These systems span across fields such as quantum mechanics, fluid dynamics, and electromagnetism. In the modern realms of science and engineering, optimizing system performance governed by physical laws is a common task across disciplines, including image processing~\cite{de2013image,kim2024numerical}, shape optimization~\cite{sokolowski1992introduction}, drug transport~\cite{chakrabarty2005optimal}, and finance~\cite{cornuejols2018optimization}. We are witnessing an increasing application of PDEs in predicting capabilities from structural analysis of high-rise buildings and tunnels to the design of cars and rockets, and even to the thermal management and electromagnetic interference shielding in the latest smartphones. Specifically, we investigate PDEs of the following form:
\begin{equation}
	\begin{cases}
		u_t + A[u] = 0, & x \in \Omega, t \in [0, T] \\
		u(x,0) = h(x), & x \in \Omega \\
		u(x,t) = g(x,t), & x \in \partial\Omega
	\end{cases}
\end{equation}
where \( u \) is the solution to the PDEs, \( A[u] \) is an operator acting on \( u \), which can be linear or nonlinear. \( \Omega \) is the domain, typically a subset of \( \mathbb{R}^D \). \( T \) is the termination time, \( h(x) \) is the initial condition that defines the state of the solution at \( t = 0 \) at position \( x \), and \( g(x,t) \) is the boundary condition that prescribes the value of the solution at the domain boundary \( \partial\Omega \) at each time point.

Over the past few decades, various numerical methods such as finite difference, finite element, and spectral methods have gradually replaced analytical approximations for linear and coupled items. These numerical methods provide effective tools for addressing complex and atypical PDEs systems. However, traditional numerical methods face significant challenges when dealing with nonlinear features, multi-scale characteristics, uncertain boundary conditions, and the complexity of high-dimensional data processing. The revolutionary progress in machine learning offers a new approach to PDEs solving, with increasing applications in solving partial differential equations. Particularly, the idea of learning a computationally cheap but sufficiently accurate substitute for classical solvers has proved very effective. Neural networks, as powerful function approximators, have been introduced into PDEs solving~\cite{greydanus2019hamiltonian,bar2019learning,raissi2019physics,li2020fourier,lu2021learning}, demonstrating substantial potential in handling complex problems. These neural network PDE solvers are laying the groundwork and rapidly becoming a fast-growing and impactful research area.

In the field of scientific machine learning (SciML), particularly in deep learning, where large volumes of data are typically required, data resources are often scarce and costly. Faced with these challenges, data augmentation~\cite{shorten2019survey,gong2024exploring} has become a cost-effective strategy for expanding training datasets~\cite{wen2020time,gong2024beyond,gong2021eliminate}. This approach not only increases the model's exposure to diverse data features but also acts as a regularization technique, helping to reduce overfitting to noise and atypical features. Although research on data augmentation in SciML is gradually increasing, the literature remains sparse, with only a few preliminary studies exploring innovative methods.

In the field of neural PDEs, current data augmentation techniques primarily rely on principles of symmetry and invariance~\cite{wang2021incorporating2}. A representative method based on symmetry, Brandstetter et al.~\cite{brandstetter2022lie} utilize Lie point symmetry for data augmentation. Lie point symmetry~\cite{eglit1996continuum} is a concept in mathematics and physics that involves exploiting the symmetry of a system to generate new solutions, thereby expanding the solution space. By leveraging this symmetry in solving PDEs, the diversity of the training dataset can be increased, improving the generalization ability of the neural network model. However, the effectiveness of the Lie point symmetry method depends on the PDEs having identifiable symmetries. Not all PDEs exhibit sufficient or apparent symmetry for this method to be viable, which limits its universality.
A representative method based on invariance, Fanaskov \cite{fanaskov2023general} employs generalized covariance for data augmentation. Generalized covariance refers to the property that physical laws remain invariant under different coordinate systems or frames of reference. This method enhances data diversity by generating different perspectives of data instances through transformations of the PDE coordinate system, thereby training the neural network to better understand and adapt to the fundamental laws and structural changes of the physical system. The effectiveness of this method is highly contingent upon the PDE system's ability to maintain its form invariant across different coordinate descriptions, which is a strong assumption for many practical problems.

Different from traditional data augmentation methods, this paper explores the application of adversarial learning in PDE solvers, specifically addressing data scarcity and enhancing practicality. By targeting and rectifying underfitted areas within the model, this approach not only effectively augments limited data resources but also substantially enhances the model's robustness and generalization capabilities. Adversarial samples are designed to reveal and exploit the vulnerabilities of models by applying meticulously designed minute perturbations to the input data, causing model predictions to deviate from true values~\cite{goodfellow2014explaining,szegedy2013intriguing,madry2018towards,gong2024crossakhtar2018threat}. Traditionally, adversarial samples have been widely used in domains such as image classification and image retrieval to evaluate and improve the robustness of models.~\cite{kurakin2018adversarial,gong2022person1,gong2024beyond2}. 

Interestingly, our research shows that adversarial samples can also play a unique role in the field of physical modeling, particularly in revealing the vulnerabilities of models when fitting physical problems. In this paper, we extend this concept to the domain of physical modeling, especially for neural network models used to solve PDEs. By incorporating adversarial samples, our method can effectively simulate various disturbances and uncertainties that might be encountered in real physical systems, and force the model to maintain stable predictive performance under these disturbances, thereby expanding the model's application range and improving its adaptability to unknown situations. 

In this research, we employ neural network-based PDE solvers, focusing on using adversarial learning strategies to enhance their robustness and generalization capabilities in scenarios characterized by data sparsity. We have developed an efficient model training strategy that allows the solver to adapt to complex partial differential equations even under conditions of data scarcity. By introducing carefully designed adversarial samples during the training process, our method effectively expands the learning range of the model and enhances its ability to handle uncertainties and potential anomalies. The effectiveness of this method is validated through experiments in various PDEs application scenarios, including equations with different physical and mathematical characteristics. In many test scenarios, the proposed method performed excellently, demonstrating significant improvements in prediction accuracy over traditional methods. These results showcase the potential of deep learning in the field of traditional scientific computing and provide empirical evidence for deploying similar technologies across a broader range of engineering and technological applications.

Our contributions are summarized as follows:

$\bullet$ Our work is the first to discuss how to design adversarial sample generation strategies tailored for the domain of physical modeling, and theoretically analyzes how the proposed method can effectively reduce overall generalization errors by integrating adversarial samples into the training process.

$\bullet$ By introducing an adversarial learning strategy, we have proposed an enhanced neural network PDEs solver that significantly improves the model's generalization capability and robustness under conditions of data scarcity and complex physical problems.

$\bullet$ Extensive experimental validation shows that our model significantly outperforms traditional methods in various complex PDEs scenarios, demonstrating the application potential of deep learning in traditional scientific computing.

\section{PDE Data Augmentation Examples}

\subsection{General Covariance Data Augmentation}
This method increases the diversity and coverage of the training dataset through coordinate transformations. Consider a simple one-dimensional elliptic equation problem:
\begin{equation}
	\begin{aligned}
		\frac{d}{dx}\left(a(x) \frac{du(x)}{dx}\right) = f(x), \\
		\quad x \in [0, 1], u(0) = u(1) = 0.
	\end{aligned}
\end{equation}

where $a(x)$ and $f(x)$ are known functions representing the coefficients and source terms of the equation.

\textbf{Coordinate Transformation Enhancement:} To apply data augmentation, a simple coordinate transformation such as $y(\xi) = \xi^3$ is chosen, which is a monotonic function from $[0, 1]$ to $[0, 1]$ satisfying $y(0) = 0$ and $y(1) = 1$.

Under the new coordinate system $\xi$, the original PDE transforms into:
\begin{equation}
	\frac{d}{d\xi}\left(a(y(\xi)) \frac{dy(\xi)}{d\xi} \frac{du(y(\xi))}{d\xi}\right) = f(y(\xi)) \left(\frac{dy(\xi)}{d\xi}\right)^2,
\end{equation}
where $\frac{dy(\xi)}{d\xi} = 3\xi^2$. With the specific transformation, the equation becomes:
\begin{equation}
	\frac{d}{d\xi}\left(a(\xi^3) 3\xi^2 \frac{du(\xi^3)}{d\xi}\right) = f(\xi^3) (3\xi^2)^2.
\end{equation}

Thus, we can generate new input-output pairs through the original solution $u(x)$ and the transformed solution $u(y(\xi))$. These augmented data will be used to train the neural network, improving its generalization ability and prediction accuracy of PDEs solutions.

\subsection{Lie Point Symmetry Data Augmentation}
Lie point symmetry is a core concept in mathematics and physics, involving identifying symmetries of partial differential equations (PDEs) that preserve solutions. By determining all possible Lie point symmetries of a PDE, we can discover multiple transformations that do not alter the fundamental structure of the equation. This method allows us to generate new solutions from known ones, thereby expanding the training dataset's size and diversity. These newly generated solutions are mathematically valid and do not require additional costly physical simulations. A crucial preparatory step before implementing Lie point symmetry-based data augmentation is to derive all the Lie point symmetry transformations associated with a specific PDE. This step is vital as it determines the types and ranges of symmetry transformations that can be applied for data augmentation.

Here, using the Korteweg-de Vries (KdV) equation as an example, we show how to generate data augmentation samples using Lie point symmetry. The KdV equation describes a single scalar field $u$ varying over space $x$ and time $t$ with the equation:
\begin{equation}
u_t + uu_x + u_{xxx} = 0,
\end{equation}
where $u_t$ is the first derivative of $u$ with respect to time, $uu_x$ is the product of $u$ and its first derivative with respect to space, and $u_{xxx}$ is the third derivative of $u$ with respect to space.

Lie point symmetry enhances the dataset through the following transformations:

\begin{enumerate}
	\item \textbf{Time Translation} $g_1(\epsilon)$:
	\begin{equation}
		g_1(\epsilon)(x, t, u) = (x, t + \epsilon, u),
	\end{equation}
	This transformation shifts the solution along the time axis by $\epsilon$.
	
	\item \textbf{Space Translation} $g_2(\epsilon)$:
	\begin{equation}
		g_2(\epsilon)(x, t, u) = (x + \epsilon, t, u),
	\end{equation}
	This transformation shifts the solution along the spatial axis by $\epsilon$.
	
	\item \textbf{Galilean Transformation} $g_3(\epsilon)$:
	\begin{equation}
		g_3(\epsilon)(x, t, u) = (x + \epsilon t, t, u + \epsilon),
	\end{equation}
	This transformation involves dynamic adjustments in both space and the solution itself.
	
	\item \textbf{Scaling Transformation} $g_4(\epsilon)$:
	\begin{equation}
		g_4(\epsilon)(x, t, u) = (e^\epsilon x, e^{3\epsilon} t, e^{-2\epsilon} u),
	\end{equation}
	This transformation adjusts the scales of space, time, and the solution.
\end{enumerate}

When training the neural network solver, by randomly selecting one or more transformation parameters $\epsilon$, we start with a solution $u$ from the training set and apply the above transformations sequentially:
\begin{equation}
	u' = g_4(\epsilon_4) g_3(\epsilon_3) g_2(\epsilon_2) g_1(\epsilon_1) u.
\end{equation}
In this way, the newly generated solutions $u'$ not only expand the size of the dataset but also enhance the model's understanding of the dynamics of physical systems and its generalization capabilities. This method requires precise symmetry derivation of the PDEs being processed before augmentation can be applied. Symmetries identified for specific PDEs may not apply to others, necessitating individual symmetry analysis and validation for each new PDEs problem.

\section{Proposed Methods}
Here we demonstrate our method using examples of a one-dimensional heat conduction equation and a two-dimensional incompressible Navier-Stokes equation.

\subsection{One-dimensional Burgers’ Equation Adversarial Sample Generation Example}

We use the one-dimensional Burgers’ equation as an example to describe how to generate adversarial samples for PDE equations, described as follows:

\begin{equation}
	\frac{\partial u}{\partial t} + u \frac{\partial u}{\partial x} = \nu \frac{\partial^2 u}{\partial x^2},
\end{equation}
where \( u(x,t) \) represents the velocity field at position \( x \) and time \( t \), and \( \nu \) is the kinematic viscosity. The model \( f(x,t;\theta) \) approximates the solution \( u(x,t) \). The loss function \( L \) is defined as:
\begin{equation}
	\begin{aligned}
		L(f(x,t;\theta), u(x,t)) = & \|f(x,t;\theta) - u(x,t)\|^2 .
	\end{aligned}
\end{equation}

The goal of generating adversarial samples is to maximize the loss function through minimal perturbations to the input. 

\subsection{Adversarial Sample Generation Steps}

\textbf{Initial Gradient Calculation.} For a given initial input data point \( (x,t) \), compute the gradient of the loss function with respect to the inputs:
\begin{equation}
	\nabla_{x,t} L(f(x,t;\theta), u(x,t)) = \left( \frac{\partial L}{\partial x} \right).
\end{equation}

\textbf{Adversarial Sample Initialization.} Set the initial adversarial sample as the original input data point:
\begin{equation}
	(x_{\text{adv}}^0, t) = (x,t).
\end{equation}

\textbf{Iterative Update of Adversarial Samples.} Gradually generate adversarial samples through multiple small-step iterations. In the \( k \)-th iteration, the adversarial sample updates as follows:
\begin{equation}
	\begin{aligned}
		(x_{\text{adv}}^{k+1}, t) = & (x_{\text{adv}}^k, t) \\
		& + \alpha \cdot \text{sign}(\nabla_x L(f(x_{\text{adv}}^k, t; \theta), u(x,t))),
	\end{aligned}
\end{equation}
where \( \alpha \) is the step size parameter, and \(\text{sign}(\cdot)\) operation determines the sign of the gradient to maximize the loss function.

\textbf{Physical Reasonableness Check.}
The goal of the physical reasonableness check is to ensure that the perturbation is within a reasonable range and meets boundary conditions. For convenience, let \( \delta_k \) represent the perturbation obtained in the \( k \)-th iteration:
\begin{equation}
	\delta_k = \alpha \cdot \text{sign}(\nabla_x L(f(x_{\text{adv}}^k, t; \theta), u(x,t))).
\end{equation}

We implement the physical reasonableness check by clipping the perturbation, expressed with the following formula:
\begin{equation}
	\begin{aligned}
		(x_{\text{adv}}^{k+1}, t) = & \ \text{clip}((x_{\text{adv}}^k, t) + \delta_k, \\
		& (-\epsilon, \epsilon), (x_{\text{min}}, t_{\text{min}}), (x_{\text{max}}, t_{\text{max}})),
	\end{aligned}
\end{equation}
where \((- \epsilon, \epsilon)\) and \((x_{\text{min}}, t_{\text{min}}), (x_{\text{max}}, t_{\text{max}})\) are respectively the clipping ranges for the perturbation amplitude and the boundaries for the physical quantities.
\((- \epsilon, \epsilon)\) ensures that the perturbation is clipped, preventing it from exceeding the maximum allowable magnitude \( \epsilon \). If \((x_{\text{adv}}^{k+1})\) exceeds \( \epsilon \) or \(- \epsilon\) after the perturbation, the clipping operation ensures they are constrained within this range. \((x_{\text{min}}, t_{\text{min}})\) and \((x_{\text{max}}, t_{\text{max}})\) are the upper and lower boundaries of the physical quantities, ensuring that the generated adversarial samples remain within a reasonable physical range. This clipping ensures that the model does not output physically unreasonable values during the generation of adversarial samples.

In traditional numerical methods, the grid granularity is the smallest unit for spatial and temporal division when discretizing a PDE, and the perturbation size \( \epsilon \) can be determined based on the grid granularity of traditional numerical methods, ensuring that the perturbation is within a reasonable range while maintaining physical consistency. Assuming that during the discretization process, the spatial variable \( x \) has a grid granularity \( \Delta x \), then the perturbation size \( \epsilon \) can be set to a value proportional to the grid granularity:
\begin{equation}
	\epsilon = \kappa \Delta x,
\end{equation}
where \( \kappa \) is a proportion coefficient less than 1, depending on the sensitivity of the physical problem and the robustness requirements of the model, ensuring that the perturbation does not exceed the grid resolution.

By matching the perturbation size with the grid granularity, the generation process of adversarial samples can be better controlled, ensuring the perturbations are physically reasonable and that the model remains stable in response to these perturbations.

\paragraph{Final Adversarial Sample Generation.}
After \( k \) iterations, the final adversarial sample is obtained:
\begin{equation}
	(x_{\text{adv}}, t) = (x_{\text{adv}}^k, t).
\end{equation}

Through the above steps and formulas, it is ensured that the generated adversarial samples are physically reasonable and do not exceed the preset boundary conditions.

\subsection{Two-dimensional Incompressible Navier-Stokes Equation Adversarial Sample Generation Example}

We use the two-dimensional incompressible Navier-Stokes equation as an example to describe how to generate adversarial samples for PDE equations, described as follows:
\begin{equation}
	\frac{\partial u}{\partial t} + u \cdot \nabla u = -\nabla p + \nu \nabla^2 u, \quad \nabla \cdot u = 0,
\end{equation}
where:

\begin{itemize}
	\item \( u = (u, v) \) is the velocity field of the fluid, with \( u \) and \( v \) representing the velocity components in the \( x \) and \( y \) directions, respectively.
	\item \( p \) is the pressure field of the fluid.
	\item \( \nu \) is the viscosity of the fluid.
\end{itemize}

The neural network model \( f(x, y, t; \theta) \) is used to approximate the solution of the velocity and pressure fields. The loss function \( L \) is defined as::
\begin{equation}
	\begin{aligned}
		L(f(x, y, t; \theta)) = & \ \|u_{\text{pred}} - u_{\text{true}}\|^2 \\
		& + \|p_{\text{pred}} - p_{\text{true}}\|^2 ,
	\end{aligned}
\end{equation}
where:
\begin{itemize}
	\item \( \|u_{\text{pred}} - u_{\text{true}}\|^2 \) represents the squared error between the predicted and true velocity fields.
	\item \( \|p_{\text{pred}} - p_{\text{true}}\|^2 \) represents the squared error between the predicted and true pressure fields.
\end{itemize}

\subsection{Adversarial Sample Generation Steps}

During the process of generating adversarial attacks for PDEs, different physical quantities might have different magnitudes, so applying the same magnitude of perturbation could lead to overly large perturbations for some quantities and too small for others. To address this issue, we introduce a normalization step, which normalizes different physical quantities to unify their dimensions. This aims to ensure a more uniform and regulated application of perturbations, avoiding inconsistencies due to differences in physical quantity scales.

Specifically, normalization converts the values of different physical quantities into a dimensionless standardized form, aligning them within the same numerical range. Using this method, we can apply a uniform perturbation step \( \alpha \) during the generation of adversarial samples without needing to adjust the perturbation magnitude for each physical quantity individually. For this purpose, each physical quantity \( q \) is normalized to obtain its dimensionless standardized form \( q_{\text{norm}} \):
\begin{equation}
	q_{\text{norm}} = \frac{q - q_{\text{min}}}{q_{\text{max}} - q_{\text{min}}},
\end{equation}
where \( q_{\text{min}} \) and \( q_{\text{max}} \) are the minimum and maximum values of the physical quantity \( q \). This transformation ensures that all physical quantities are normalized within the range [0, 1].

\textbf{Initial Gradient Calculation.} For a given initial input data point \( (x, y, t) \), compute the gradient of the loss function with respect to the spatial variables:
\begin{equation}
	\nabla_{x, y} L(f(x, y, t; \theta)) = \left( \frac{\partial L}{\partial x}, \frac{\partial L}{\partial y} \right).
\end{equation}

\textbf{Adversarial Sample Initialization:} Set the initial adversarial sample as the original input data point:
\begin{equation}
	(x_{\text{adv}}^0, y_{\text{adv}}^0, t) = (x, y, t).
\end{equation}

\textbf{Iterative Update of Adversarial Samples.} Gradually generate adversarial samples through multiple small-step iterations. In the \( k \)-th iteration, the adversarial sample updates as follows:
\begin{equation}
	\begin{aligned}
		(x_{\text{adv}}^{k+1}, y_{\text{adv}}^{k+1}, t) = & (x_{\text{adv}}^k, y_{\text{adv}}^k, t) \\
		& + \alpha \cdot \text{sign}(\nabla_{x, y} L(f(x_{\text{adv}}^k, y_{\text{adv}}^k, t; \theta))),
	\end{aligned}
\end{equation}
where \( \alpha \) is the step size parameter, and \(\text{sign}(\cdot)\) operation determines the sign of the gradient to maximize the loss function.

\textbf{Physical Reasonableness Check.} After each update, the adversarial sample undergoes a physical reasonableness check to ensure that the perturbation is within a reasonable range and does not disrupt the continuity of time-series data. Specifically, perturbations should only be applied in the spatial dimensions, not involving the time dimension. The size of the perturbation \( \epsilon \) can be determined based on the grid granularity to ensure that the generated adversarial sample is physically reasonable:
\begin{equation}
	\begin{aligned}
		(x_{\text{adv}}^{k+1}, y_{\text{adv}}^{k+1}, t) = & \ \text{clip}((x_{\text{adv}}^k, y_{\text{adv}}^k, t) + \delta_k, \\
		& (x_{\text{min}}, y_{\text{min}}, t_{\text{min}}), (x_{\text{max}}, y_{\text{max}}, t_{\text{max}})),
	\end{aligned}
\end{equation}
where:
\begin{equation}
	\delta_k = \alpha \cdot \text{sign}(\nabla_{x, y} L),
\end{equation}
represents the perturbation obtained in the \( k \)-th iteration.

\textbf{Final Adversarial Sample Generation.} After \( k \) iterations, the final adversarial sample is obtained:
\begin{equation}
	(x_{\text{adv}}, y_{\text{adv}}, t) = (x_{\text{adv}}^k, y_{\text{adv}}^k, t).
\end{equation}

Through these steps, adversarial samples that are physically reasonable can be generated, maintaining the continuity and reasonableness of time-series data. These adversarial samples will be used to test the robustness and consistency of the model, ensuring that the model can effectively cope with complex physical conditions in real-world applications.

\subsection{Theoretical Analysis}
The purpose of generating adversarial samples \(S_{adv}\) is to challenge the model by exposing it to regions in the input space where its predictions may be weak. By identifying and strengthening the model's fitting ability in these critical regions, we aim to reduce the generalization error.

We introduce a coverage measure \(C(f_\theta, S)\), which represents the total error of the model \(f_\theta\) over the entire data distribution \(S\). For the original data distribution \(S\), the coverage measure is defined as:
\begin{equation}
C(f_\theta, S) = \int_S \|f_\theta(x) - u(x)\|^2 \, dx,
\end{equation}
where \(f_\theta(x)\) represents the model's prediction and \(u(x)\) represents the true solution or target value.

After introducing adversarial samples \(S_{adv}\), which are perturbations of the original data points, the new coverage measure can be expressed as:
\begin{equation}
C(f_\theta, S \cup S_{adv}) = \int_{S \cup S_{adv}} \|f_\theta(x) - u(x)\|^2 \, dx.
\end{equation}

By including adversarial samples, \(S_{adv}\), the coverage measure now accounts for potential vulnerabilities in the model, and the errors associated with these adversarial samples are explicitly minimized during training. This leads to an overall reduction in the error:
\begin{equation}
C(f_\theta, S) \geq C(f_\theta, S \cup S_{adv}).
\end{equation}

This indicates that adversarial training reduces the model's generalization error.

\begin{figure*}[t]
	\setlength{\abovecaptionskip}{0.1cm}
	\setlength{\belowcaptionskip}{-0.4cm}   
	\centering
	\includegraphics[width=1\linewidth]{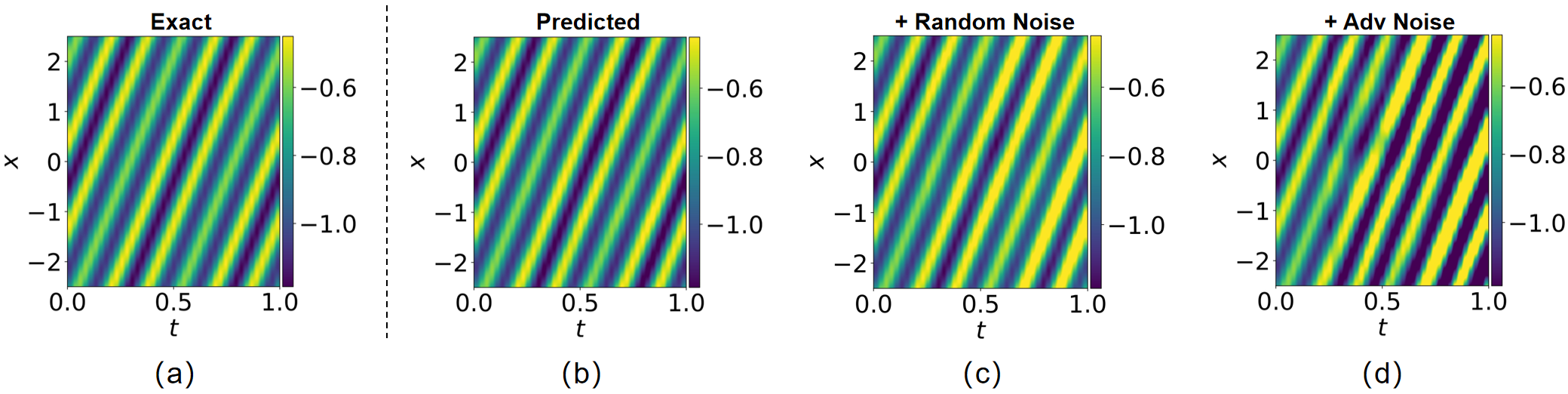}
	\caption{Displays the time evolution of the 1D Advection equation: (a) exact solution, (b) neural network prediction, (c) prediction with added random noise, (d) prediction after adversarial perturbation. The amplitude of both random and adversarial noise is 8\% of the grid size.}
	\label{pic1}
\end{figure*}

\subsection{Example Pseudocode}
The supplementary materials provide the pseudocode for the proposed SMART method. This approach systematically generates adversarial samples to challenge the model's performance in areas where predictions may be vulnerable. Initially, the model generates predictions based on the input data and calculates the loss function. Then, slight perturbations are applied to the input data using the gradient of the loss, creating adversarial samples that are used for further training. By incorporating these adversarial samples during training, the model's generalization ability is enhanced, effectively reducing overall generalization error. For more details, please refer to the supplementary materials.

\section{Experiments}

\begin{table*}[]\small
	\centering
	\caption{Performance Comparison of Standard Training (Standard) and Ours on Different Numbers of Training Points for the 1D Burgers and 1D Advection Equations. RMSE = Root Mean Square Error, N RMSE = normalized RMSE, RMSE C = RMSE of conserved variables, RMSE B = RMSE at boundaries, Max error = Maximum value of rms error. The gain is calculated as $g = \left( 1 - \frac{E_{\text{Ours}}}{E_{\text{Standard}}}\right) \times 100\%$.}
	\setlength{\tabcolsep}{5pt} 
	\begin{tabular*}{\textwidth}{@{\extracolsep{\fill}} l l c c c c c c c c c c c @{}}
		\toprule
		\multirow{2}{*}{Num} & \multirow{2}{*}{Training} & \multicolumn{5}{c}{1D Burgers equation} & \multicolumn{5}{c}{1D Advection equation} \\ \cmidrule(lr){3-7} \cmidrule(lr){8-12}
		&  & RMSE & N RMSE & RMSE C & RMSE B & Max Error & RMSE & N RMSE & RMSE C & RMSE B & Max Error \\ \midrule
		\multirow{3}{*}{32} 
		& Standard & 0.00506 & 0.00733 & 0.00884 & 0.00928 & 0.05485 & 0.12211 & 0.15733 & 0.01988 & 0.14811 & 0.76582 \\ 
		& Ours & 0.00284 & 0.00607 & 0.00262 & 0.00692 & 0.02714 & 0.09523 & 0.12586 & 0.01591 & 0.11734 & 0.62528 \\ 
		& $g$ & 43.87\% & 17.19\% & 70.36\% & 25.43\% & 50.52\% & 22.02\% & 20.01\% & 19.98\% & 20.76\% & 18.37\% \\ \midrule
		
		\multirow{3}{*}{64}
		& Standard & 0.00439 & 0.01057 & 0.00592 & 0.00836 & 0.01740 & 0.08974 & 0.11257 & 0.01456 & 0.11436 & 0.63521 \\
		& Ours & 0.00234 & 0.00658 & 0.00258 & 0.00329 & 0.01131 & 0.07391 & 0.09472 & 0.00947 & 0.09321 & 0.53417 \\
		& $g$ & 46.70\% & 37.75\% & 56.42\% & 60.65\% & 35.00\% & 17.63\% & 15.83\% & 34.97\% & 18.48\% & 15.91\% \\ \midrule
		
		\multirow{3}{*}{128}
		& Standard & 0.00204 & 0.00915 & 0.00045 & 0.01179 & 0.01523 & 0.06132 & 0.08745 & 0.00935 & 0.08578 & 0.51345 \\
		& Ours & 0.00115 & 0.00465 & 0.00034 & 0.00408 & 0.00688 & 0.04452 & 0.05939 & 0.00748 & 0.05842 & 0.42638 \\
		& $g$ & 43.63\% & 49.18\% & 24.44\% & 65.39\% & 54.83\% & 27.38\% & 32.09\% & 20.00\% & 31.88\% & 16.96\% \\ \midrule
		
		\multirow{3}{*}{256}
		& Standard & 0.00147 & 0.00808 & 0.00027 & 0.00507 & 0.01103 & 0.03684 & 0.06348 & 0.00576 & 0.06237 & 0.37602 \\
		& Ours & 0.00126 & 0.00673 & 0.00025 & 0.00458 & 0.00927 & 0.03212 & 0.05471 & 0.00412 & 0.05298 & 0.30321 \\
		& $g$ & 14.29\% & 16.71\% & 7.41\% & 9.66\% & 15.96\% & 12.82\% & 13.81\% & 28.47\% & 15.06\% & 19.36\% \\ \midrule
		
		\multirow{3}{*}{512}
		& Standard & 0.00087 & 0.00418 & 0.00028 & 0.00158 & 0.00846 & 0.01323 & 0.02491 & 0.00200 & 0.02194 & 0.19004 \\
		& Ours & 0.00082 & 0.00392 & 0.00027 & 0.00145 & 0.00712 & 0.01234 & 0.02342 & 0.00175 & 0.02150 & 0.18239 \\
		& $g$ & 5.75\% & 6.22\% & 3.57\% & 8.23\% & 15.84\% & 6.73\%  & 5.97\%  & 12.50\% & 2.00\%  & 4.03\% \\ \bottomrule
	\end{tabular*}
	\label{tab1}
\end{table*}

\begin{table*}[h]
	\centering
	\caption{Comparison of General Covariance Data Augmentation (GCDA) and our SMART Methods on PDE Solving: 2D Navier-Stokes Equation. The values in parentheses represent the corresponding percentage gain.}
	\setlength{\tabcolsep}{6pt} 
	\renewcommand{\arraystretch}{1.2} 
	\begin{tabular}{cccccccc}
		\toprule
		\textbf{Equation} & \textbf{Model} & \multicolumn{3}{c}{\textbf{v1}} & \multicolumn{3}{c}{\textbf{v2}} \\
		\cmidrule(lr{0pt}){3-5} \cmidrule(lr{0pt}){6-8}
		& & \textbf{Error} & \textbf{GCDA} & \textbf{Ours} & \textbf{Error} & \textbf{GCDA} & \textbf{Ours} \\
		\midrule
		\multirow{4}{*}{\centering Navier-Stokes} & \centering FNO 
		& \centering 0.007 & 0.005 (28.57\%) & \textbf{0.003 (57.14\%)} & 0.024 & 0.016 (33.33\%) & \textbf{0.012 (50.00\%)} \\
		& \centering DilResNet & \centering 0.023 & 0.018 (21.74\%) & \textbf{0.012 (47.83\%)} & 0.071 & 0.052 (26.76\%) & \textbf{0.046 (35.21\%)} \\
		& \centering MLP & \centering 0.088 & 0.069 (21.59\%) & \textbf{0.060 (31.82\%)} & 0.086 & 0.074 (13.95\%) & \textbf{0.062 (27.91\%)} \\
		& \centering SNO & \centering 0.005 & 0.004 (20.00\%) & \textbf{0.004 (40.00\%)} & 0.014 & 0.011 (21.43\%) & \textbf{0.007 (50.00\%)} \\
		\bottomrule
	\end{tabular}
	\label{tab2}
\end{table*}
	
As shown in Fig.~\ref{pic1}, by comparing panels (b), (c), and (d), one can visually observe the impact of adversarial perturbations on the accuracy of the model's predicted solutions. Both random noise and adversarial noise were set to 8\% of the grid size. Compared to random noise, the impact of adversarial noise on the model's predictions is significantly more pronounced. The effects of random and adversarial noise on the model across various metrics are presented in Tab. 1 of the supplementary materials.

These results demonstrate that small adversarial perturbations can significantly degrade the model's predictive accuracy, revealing vulnerabilities at specific data points. These adversarial perturbations, which form adversarial examples, are crucial for further optimizing and enhancing the model's robustness. 

As shown in Supplementary Material Fig. 1, compared to traditional standard training methods, our SMART training strategy proposed in this paper achieves a faster decrease in training loss across datasets with different numbers of training points, especially during the initial stages of training. Our SMART strategy requires fewer iterations to converge in all training point configurations, thereby demonstrating its superiority in both efficiency and effectiveness.

\subsection{Evaluation Criteria}
The experiments evaluate several metrics~\cite{2022pdebench}, including RMSE, normalized RMSE (N RMSE), RMSE of conserved variables (RMSE C), RMSE at boundaries (RMSE B), and Max Error. These metrics provide insights into the accuracy and physical consistency of the model's predictions. Lower values in these metrics generally indicate higher prediction accuracy, better adherence to physical laws, and greater robustness. Additionally, we calculate data gain using \( g = \left(1 - \frac{E_{\text{method}}}{E_{\text{test}}}\right) \times 100\% \) to provide a more intuitive demonstration of the method's effectiveness, where \(E_{\text{method}}\) is the test error of the proposed method, and \(E_{\text{test}}\) is the test error of the comparison target.

\subsection{Comparison Experiments}


\begin{table*}[]
	\vspace{0.3cm} 
	\centering
	\caption{Percentage Gain on the Burgers' Equation (with 10 time steps) using Lie Point Symmetry Data Augmentation (LPSDA) combined with our method. The symmetry transformations \(g1\) and \(g5\) correspond to time translation and scaling, respectively, with \(g1g5\) denoting the combined application of both transformations. FNO (AR) and FNO (NO) respectively denote the Autoregressive and Neural Operator training methods applied to the Fourier Neural Operator (FNO) model. The percentage gain is calculated based on the formula \( (1 - \frac{E_{\text{aug}}}{E_{\text{test}}}) \times 100\% \). Positive gain values indicate a performance improvement.}
	\label{tab:burgers_lpsda_ours_gain}
	\renewcommand{\arraystretch}{1.2} 
	\setlength{\tabcolsep}{10pt} 
	\begin{tabular}{@{}cccp{1.4cm} p{1.1cm}p{1.1cm}p{1.1cm}p{1.1cm}p{1.4cm}@{}}
		\toprule
		\textbf{Task} & \textbf{Solver} & \textbf{Symmetry} &\textbf{Method} & \textbf{512} & \textbf{256} & \textbf{128} & \textbf{64} & \textbf{32} \\
		\midrule
		\multirow{8}{*}{Burgers} & \multirow{2}{*}{FNO(AR)} & \multirow{2}{*}{g1} & LPSDA & \mbox{50.0\%} & \mbox{32.97\%} & \mbox{34.63\%} & \mbox{28.12\%} & \mbox{22.45\%} \\
		&  &  & + Ours & \mbox{+5.0\%} & \mbox{+9.89\%} & \mbox{+10.2\%} & \mbox{+9.33\%} & \mbox{+8.47\%} \\
		\cmidrule{3-9}
		& \multirow{2}{*}{FNO(AR)} & \multirow{2}{*}{g1g5} & LPSDA & \mbox{85.0\%} & \mbox{89.01\%} & \mbox{90.95\%} & \mbox{78.23\%} & \mbox{66.42\%} \\
		&  &  & + Ours & \mbox{+5.0\%} & \mbox{+3.30\%} & \mbox{+1.15\%} & \mbox{+7.21\%} & \mbox{+7.13\%} \\
		\cmidrule{2-9}
		& \multirow{2}{*}{FNO(NO)} & \multirow{2}{*}{g1} & LPSDA & \mbox{78.50\%} & \mbox{50.55\%} & \mbox{2.47\%} & \mbox{-10.12\%} & \mbox{-20.34\%} \\
		&  &  & + Ours & \mbox{+2.11\%} & \mbox{+5.97\%} & \mbox{+43.31\%} & \mbox{+52.76\%} & \mbox{+57.76\%} \\
		\cmidrule{3-9}
		& \multirow{2}{*}{FNO(NO)} & \multirow{2}{*}{g1g5} & LPSDA & \mbox{99.30\%} & \mbox{99.21\%} & \mbox{97.64\%} & \mbox{85.01\%} & \mbox{72.44\%} \\
		&  &  & + Ours & \mbox{+0.23\%} & \mbox{+0.18\%} & \mbox{+0.47\%} & \mbox{+2.31\%} & \mbox{+10.21\%} \\
		\bottomrule
	\end{tabular}
\end{table*}

In our experiments on the 1D Burgers equation, Tab.~\ref{tab1} presents the performance comparison between the standard training method and our proposed approach across different numbers of training points. The results indicate that our method demonstrates significant advantages in scenarios with limited training data (sparse data scenarios). As the number of training points increases, the overall accuracy of the model improves, leading to a substantial reduction in error values. For instance, when the number of training points is 32, our method achieves a RMSE gain of 43.87\%; however, this gain decreases to 5.75\% as the number of training points reaches 512. This trend suggests that as the model’s accuracy improves, further enhancements become more challenging, reflecting a common phenomenon in machine learning where gains diminish as performance approaches optimal levels. In the 1D Advection equation, it can be observed that the experimental results are similar to those obtained with the 1D Burgers equation. The experimental results for the 2D CFD equation can be found in Tab. 2 of the Supplementary Material. Our approach shows considerable potential for improving model performance in data-limited scenarios, while still providing robustness as data volume increases.

The results in Tab.~\ref{tab2} compare the performance of our SMART method against General Covariance Data Augmentation (GCDA) in solving two components of the fluid velocity field for the two-dimensional Navier-Stokes equation across various models, including Fourier Neural Operator (FNO)~\cite{li2020fourier}, Dilated Residual Network (DilResNet)~\cite{yu2015multi,stachenfeld2021learned}, Multilayer Perceptron (MLP)~\cite{haykin1994neural}, and Structured Neural Operator (SNO)~\cite{fanaskov2023spectral}. A particularly notable improvement is observed in the SNO model for the \(v2\) component, where the error reduction using SMART jumps from 21.43\% with GCDA to 50.00\%, more than doubling the performance gain. This significant increase clearly demonstrates the superior ability of the SMART method to enhance model accuracy, particularly in complex physical modeling scenarios. Overall, the consistent performance improvements across all models underscore the effectiveness of SMART in reducing errors and enhancing the robustness of predictions. Additional comparative experiments are provided in the supplementary materials.

Our method is complementary to existing approaches and demonstrates greater adaptability. Table~\ref{tab:burgers_lpsda_ours_gain} shows the percentage gain in solving the Burgers' equation when combining Lie Point Symmetry Data Augmentation (LPSDA) with our method. The column labeled “+ Ours” represents the additional gain achieved by integrating our method with LPSDA. Notably, when training the FNO(NO) model, the application of the g1 symmetry transformation in LPSDA resulted in a negative gain, revealing some limitations of the approach in certain scenarios. However, with the introduction of our method, this negative gain was significantly mitigated, indicating that our approach effectively compensates for the shortcomings of LPSDA and significantly enhances model performance under challenging conditions.

\section{Conclusion}
This paper introduced an innovative adversarial learning approach called Systematic Enhancement with Adversarial Robust Training (SMART), aimed at enhancing the robustness and generalization of neural network PDE solvers in sparse data scenarios. Through extensive experiments, our method demonstrated superior performance in complex PDE scenarios compared to traditional data augmentation techniques. Moreover, our theoretical analysis supports the empirical findings, indicating that adversarial learning effectively expands the exploration range of PDE solutions and reduces the model's generalization error. Future work will focus on integrating adversarial learning into broader areas of scientific computing to address challenges posed by data scarcity.

{\small
\bibliographystyle{ieee_fullname}
\bibliography{egbib}
}

\end{document}